\documentclass[letterpaper]{article} % DO NOT CHANGE THIS
\usepackage{aaai2026}

\usepackage{times}  % DO NOT CHANGE THIS
\usepackage{helvet}  % DO NOT CHANGE THIS
\usepackage{courier}  % DO NOT CHANGE THIS
\usepackage[hyphens]{url}  % DO NOT CHANGE THIS
\usepackage{graphicx} % DO NOT CHANGE THIS
\usepackage{natbib}  % DO NOT CHANGE THIS AND DO NOT ADD ANY OPTIONS TO IT
\usepackage{caption} % DO NOT CHANGE THIS AND DO NOT ADD ANY OPTIONS TO IT
\usepackage{algorithm}
\usepackage{algorithmic}

\usepackage{newfloat}
\usepackage{listings}
\usepackage{booktabs}
\usepackage{amssymb}
\usepackage{pifont}
\usepackage{multirow} 
\usepackage{amsmath} 
\usepackage[most]{tcolorbox}  % 用于美观的框框
\usepackage{enumitem}         % 控制 itemize 样式和缩进
\usepackage{xcolor}           % 颜色支持
\usepackage{verbatim}         % 多行文本格式化（如 output 示例）
\usepackage{bibentry}

\DeclareCaptionStyle{ruled}{labelfont=normalfont,labelsep=colon,strut=off} % DO NOT CHANGE THIS
\floatstyle{ruled}
\newfloat{listing}{tb}{lst}{}
\floatname{listing}{Listing}

\title{Vision‑Language Reasoning for Geolocalization: A Reinforcement Learning Approach}

\author{
 Biao Wu$^{1}$, Meng Fang$^{2}$, Ling Chen$^{1}$, Ke Xu$^{2}$, Tao Cheng$^{3}$,  Jun Wang$^{3}$  \\
}

\affiliations {
    \textsuperscript{\rm 1}University of Technology Sydney, Australia \\
    \textsuperscript{\rm 2}University of Liverpool, United Kingdom \\ 
    \textsuperscript{\rm 3}University College London, United Kingdom \\
    biao.wu-2@student.uts.edu.au, Meng.Fang@liverpool.ac.uk, Ling.Chen@uts.edu.au,  \\ 
    Ke.Xu@liverpool.ac.uk, tao.cheng@ucl.ac.uk, jun.wang@ucl.ac.uk
}

\begin{document}

\maketitle

\begin{abstract}
Recent advances in vision-language models have opened up new possibilities for reasoning-driven image geolocalization. However, existing approaches often rely on synthetic reasoning annotations or external image retrieval, which can limit interpretability and generalizability. In this paper, we present Geo-R, a retrieval-free framework that uncovers structured reasoning paths from existing ground-truth coordinates and optimizes geolocation accuracy via reinforcement learning. We propose the Chain of Region, a rule-based hierarchical reasoning paradigm that generates precise, interpretable supervision by mapping GPS coordinates to geographic entities (e.g., country, province, city) without relying on model-generated or synthetic labels. Building on this, we introduce a lightweight reinforcement learning strategy with coordinate-aligned rewards based on Haversine distance, enabling the model to refine predictions through spatially meaningful feedback. Our approach bridges structured geographic reasoning with direct spatial supervision, yielding improved localization accuracy, stronger generalization, and more transparent inference. Experimental results across multiple benchmarks confirm the effectiveness of Geo-R, establishing a new retrieval-free paradigm for scalable and interpretable image geolocalization. To facilitate further research and ensure reproducibility, all relevant resources, including the model and code, are publicly available at \url{https://github.com/aialt/geo-r}.
\end{abstract}

% Uncomment the following to link to your code, datasets, an extended version or similar.
% You must keep this block between (not within) the abstract and the main body of the paper.
% \begin{links}
%     \link{Code}{https://aaai.org/example/code}
%     \link{Datasets}{https://aaai.org/example/datasets}
%     \link{Extended version}{https://aaai.org/example/extended-version}
% \end{links}

\section{Introduction}

\begin{figure*}[t]
\centering
\includegraphics[width=1\textwidth]{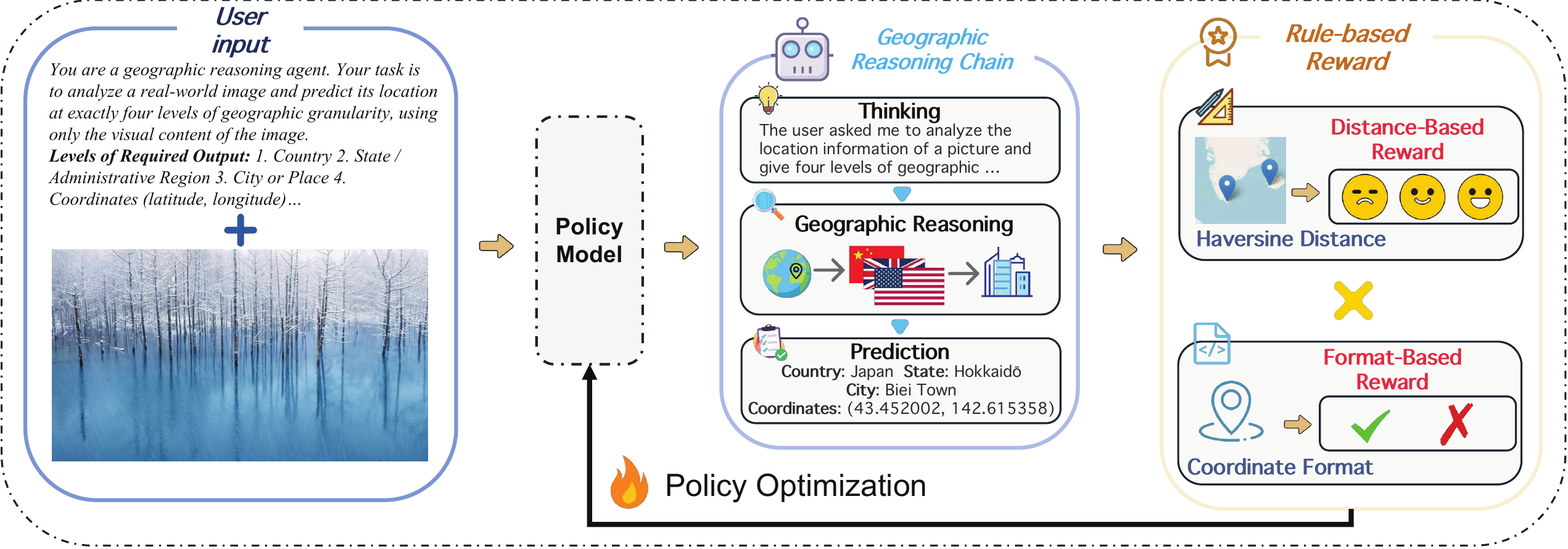}  % 0.478 ≈ 9/16
\caption{
Overview of Geo-R for image geolocation. Given a query image, the agent generates a region-level reasoning chain grounded in geographic knowledge, followed by coordinate prediction. The framework integrates structured prompting, multimodal understanding, and reinforcement optimization to enhance both interpretability and spatial accuracy.
}
\label{fig:geo_agent_overview}
% \vspace{-3mm}
\end{figure*}

Image geolocalization—the task of predicting the geographic coordinates of an image—poses unique difficulties, particularly at the global scale. The inherent difficulty of this task arises from the high diversity of geographic regions, the visual similarity between distant locations, and the lack of explicit geographic cues in many images. Existing methods mainly fall into two categories: classification-based approaches, which partition the Earth into discrete regions and treat geolocation as a classification task~\cite{weyand2016, seo2018, muller2018, pramanick2022, clark2023}, and retrieval-based approaches, which retrieve the most visually similar samples from large-scale geo-tagged image databases~\cite{yang2021cross, zhu2022transgeo, wang2023fine, vivanco2023geoclip, haas2024pigeon, xia2025fg, haas2023learning}. While both paradigms have achieved strong benchmark performance, they rely heavily on large-scale labeled data or retrieval databases and often lack interpretability and robust reasoning capabilities, limiting their generalization to unseen regions.

Reasoning for geolocalization is challenging and less studied, as it requires models to generate interpretable, structured explanations that align with visual evidence—a task further complicated by the scarcity of annotated reasoning paths. While recent advances in vision-language modeling have driven progress in related areas such as visual recognition, language grounding, and cross-modal retrieval~\cite{chen2023sharegpt4v,koh2023grounding,wang2025infinity,wu2024foundations}, there is still little work addressing geolocalization using vision-language models (VLMs). Techniques such as few-shot prompting, retrieval-augmented generation (RAG), and supervised fine-tuning (SFT)~\cite{jia2024g3, zhou2024img2loc} allow models to produce both geolocation predictions and their corresponding reasoning processes. However, these methods typically rely on synthetic data to construct structured reasoning annotations, which only partially addresses the lack of real-world annotated data. Synthetic supervision frequently results in shallow or inconsistent reasoning that may not accurately reflect the visual evidence, thus constraining overall model performance.

To address these challenges, we propose Geo-R, a retrieval-free, reasoning-centric model for global image geolocalization built on the Geo Chain-of-Thoughts (CoT) paradigm. Geo-R mimics human geographic reasoning by progressively localizing an image from country to state to city, and finally estimating its precise coordinates. By clearly labeling each geographic level, this hierarchical approach decomposes the complex geolocation task into a sequence of interpretable subgoals, reducing the difficulty of direct coordinate regression and explicitly leveraging the latent geographic priors embedded in vision–language models (VLMs), thereby improving both accuracy and interpretability. Meanwhile, we construct MP16-Rand-500K, a large-scale, high-quality geographic reasoning dataset covering diverse global locations. Its reasoning sequences are generated by reverse-decoding ground-truth coordinates into multi-level geographic information, producing a coherent coarse-to-fine structure aligned with the hierarchical prediction process of Geo-R. This dataset is employed in the SFT stage to further enhance the model’s structured geographic reasoning ability, laying a solid foundation for more advanced training and evaluation.

Furthermore, we observe that the core evaluation metric in geolocation tasks is the proximity of predicted coordinates to the ground truth. However, in regression tasks, SFT usually does not penalize small numerical errors, making it difficult to improve prediction accuracy directly. In contrast, Reinforcement Learning (RL) provides continuous and directional optimization signals for numerical errors—for example, assigning higher rewards as predicted coordinates approach the ground truth. Based on this insight, we introduce an RL stage and optimize the model within the Group Relative Policy Optimization (GRPO)~\cite{shao2024deepseekmath}, using a composite reward function that jointly accounts for spatial accuracy and faithful reasoning. However, during training, we identify a critical issue with GRPO: the vanishing advantages problem. When all responses within a query group receive identical rewards (e.g., all correct or all incorrect), the computed relative advantages approach zero, resulting in ineffective gradient updates. This issue is especially pronounced in later training stages, where a large portion of samples originate from “popular regions” such as city centers or iconic landmarks. These regions often exhibit salient visual cues and low localization difficulty, leading to uniform predictions and reward saturation—ultimately limiting the model’s capacity to generalize to more complex scenarios. To mitigate this, we introduce a diversity-based data filtering mechanism and construct a more challenging subset, MP16-Hard-200K, by excluding all samples within a 200-kilometer radius of popular regions and focusing instead on remote, visually ambiguous, or culturally neutral areas. These long-tailed and difficult examples significantly enhance training diversity and reward signal variance, effectively alleviating the vanishing advantages problem and further improving the model’s generalization and reasoning performance in complex, real-world geographic environments.

We conduct comprehensive evaluations of Geo-R on two standard geolocalization benchmarks. Experimental results demonstrate that our method achieves strong performance across both coarse-grained tasks at the country levels and fine-grained tasks at the regional level. Compared to retrieval-based baselines, Geo-R not only delivers competitive localization accuracy but also provides superior interpretability, cross-domain generalization, and robustness across a wide range of geographic conditions. These findings validate the feasibility and effectiveness of a retrieval-free, reasoning-driven paradigm for geolocalization and point toward a new direction for building scalable and general-purpose geographic agents.

\noindent
Our key contributions are fourfold:

\begin{itemize}
    \item We introduce Geo-R, a retrieval‑free, reasoning‑driven architecture for global image geolocalization.

    \item We propose the geo Chain‑of‑Thoughts paradigm and accompany it with a scalable data‑synthesis pipeline and richly annotated geographic‑reasoning dataset, enabling the model to learn structured, hierarchical inference from visual cues. 

    \item We develop a novel reinforcement‑learning framework with plausibility rewards, directly optimizing geographic accuracy via spatial‑distance–based signals while enforcing strict output formatting.

    \item Our experiments demonstrate that pure reasoning—without any external retrieval—can match or surpass traditional retrieval‑based pipelines in both localization accuracy and cross‑domain generalization across diverse global environments.

\end{itemize}

\noindent

\section{Related Work}

\subsection{Image Geolocalization}

Image geo-localization aims to predict the geographic coordinates of a given image and has wide applications in urban analysis~\cite{yeh1999urban, firmansyah2024empowering, yan2024urbanclip, ye2019measuring, ye2019visual}, navigation~\cite{desouza2002vision}, and geospatial data mining~\cite{korting2013geodma, liu2024unitime, liang2018geoman, pan2019urban, hao2025nature}. With the rapid progress of multimodal learning, the research paradigm for image geo-localization has gradually shifted from traditional classification-based methods~\cite{weyand2016, seo2018, muller2018, pramanick2022, clark2023} and retrieval-based pipelines~\cite{yang2021cross, zhu2022transgeo, wang2023fine, vivanco2023geoclip, xia2025fg, haas2023learning} to generation-based reasoning approaches~\cite{zhou2024img2loc, jia2024g3} that offer better interpretability and generalization. These methods utilize vision-language models (VLMs) to produce structured reasoning paths and coordinate predictions. Recent work has proposed synthetic reasoning datasets constructed by distilling reasoning capabilities from multiple VLMs, yielding annotations such as locatability assessments, reasoning trajectories, and predicted coordinates. Based on this, models are further optimized using Group Relative Policy Optimization (GRPO), significantly improving interpretability and geographic accuracy. However, such approaches still rely on distilled synthetic reasoning data, which may suffer from hallucinations, redundancy, or structural inconsistency. To address these limitations, we propose Geo-R, a truly retrieval-free and distillation-free reasoning-centric framework that effectively integrates structured reasoning supervision with coordinate-level optimization objectives, thereby enhancing both semantic interpretability and geographic precision with improved robustness.

\subsection{Verifiable Rewards for VLMs}

Recent advances in multimodal learning have sparked growing interest in using reinforcement learning with verifiable rewards to enhance the reasoning capabilities of VLMs. A series of studies have shown that R1-guided training can promote structured inference: R1-OneVision~\cite{yang2025r1}, Infinity Parserv~\cite{wang2025infinity} create visual reasoning datasets by translating images into textual form and apply reward-driven fine-tuning to improve interpretability. R1-V~\cite{chen2025r1v} adopts GRPO~\cite{shao2024deepseekmath}, introduced in DeepSeek R1\cite{guo2025deepseek}, to outperform much larger models in object-counting tasks. Additional work such as VisualThinker-R1-Zero~\cite{zhou2025r1} and Vl-Rethinker~\cite{wang2025vl} reveals emergent reasoning abilities like the “visual aha moment” through R1 and~\cite{kool2019buy}. Other directions include curriculum-based reward strategies~\cite{deng2025boosting} and joint text-multimodal RL training (LMM-R1~\cite{peng2025lmm}). While most of these efforts focus on mathematical reasoning~\cite{lu2023mathvista}, works like Visual-RFT~\cite{liu2025visual} begin to explore perceptual reasoning. In contrast, we apply GRPO to spatial reasoning with retrieval-free supervision, directly optimizing coordinates with geographic plausibility rewards to achieve accurate and interpretable geolocalization.

\section{Methodology}

In this section, we present Geo-R, a reasoning-centric and retrieval-free framework for global image geolocalization that bridges symbolic reasoning and spatial precision. Geo-R enables VLMs to infer geographic locations through structured, interpretable, and verifiable reasoning chains. The framework comprises three key components: (1) Chain-of-Region (CoR), a reasoning paradigm that guides the model to infer hierarchical geographic labels based on visual cues; (2) a reinforcement learning stage optimized under the GRPO framework, which uses a composite reward to jointly improve coordinate accuracy and format consistency; and (3) a diversity-driven data selection strategy, which constructs a challenging subset from visually ambiguous and underrepresented regions to enhance generalization. 
% Together, these components enable Geo-R to move beyond shallow coordinate regression or synthetic distillation, supporting robust geographic understanding grounded in multimodal reasoning.

\subsection{Synthesizing Geographic Reasoning Data at Scale}

To effectively activate the latent geographic knowledge and spatial reasoning capabilities of VLMs, we propose CoR, a structured reasoning paradigm that reformulates image geolocalization as a step-by-step inferential process. While VLMs possess extensive world knowledge and cross-modal understanding, we observe that directly regressing geographic coordinates often fails to elicit deeper geographic cognition. In contrast, VLMs are inherently better suited for structured reasoning, hierarchical decision-making, and language-driven inference chains.  Our CoR addresses this by guiding the model to generate spatial judgments based on visible cues in the image—such as landmarks, vegetation, architectural styles, and climate—thereby enabling a reasoning transition from what is seen to where it is. 

% As shown in the Prompt Template 1 in appendix,

To support this paradigm at scale, we design a data synthesis pipeline that automatically generates richly annotated samples for geographic reasoning. A key insight is that each reasoning label—country, region, city—can be automatically derived by reverse-decoding the ground-truth coordinates using global administrative boundary databases and geocoding tools. This removes the need for manual annotation while aligning perfectly with our CoR output format.  In total, we synthesize 500K geographically diverse reasoning samples, denoted as MP16-Rand-500k, forming a solid foundation for training VLMs to reason about place with precision and structure.

\subsection{Reinforcement Learning for Geographic Reasoning}

To enable reinforcement learning with verifiable rewards for geographic reasoning, we adopt GRPO as our training framework~\cite{shao2024deepseekmath}. Unlike traditional methods that evaluate each output in isolation, GRPO compares multiple candidate responses generated from the same input to estimate their relative advantages, enabling group-wise ranking to guide stable and informed policy updates. To support this framework and better align with the characteristics of geographic reasoning, we design a composite reward function that jointly captures both spatial precision and format consistency. Specifically, we incorporate two complementary components: a distance-based reward, which uses Haversine distance to measure the geodesic error between the predicted and ground-truth coordinates—encouraging geographically accurate and directionally reasonable predictions—and a format-based reward, which assigns a positive score only when the model output includes a single, well-structured, and parsable latitude-longitude pair. By combining these two components, the reward function ensures that only predictions which are both accurate and interpretable contribute to policy optimization, thereby providing fine-grained, verifiable learning signals well-suited to the demands of geographic reasoning.

\paragraph{Distance-Based Reward}  
The first component of our reward function measures the spatial accuracy of the predicted coordinates $(x_1, y_1)$ against the ground-truth coordinates $(x_2, y_2)$ using the Haversine distance~\cite{sinnott1984virtues}, which accounts for the curvature of the Earth. Let $R = 6371$ km denote the Earth's mean radius.

We first compute the differences in latitude and longitude (in radians) between the predicted and actual locations:

\begin{equation}
\Delta x = x_2 - x_1, \quad \Delta y = y_2 - y_1
\end{equation}
\vspace{1mm}

where $\Delta x$ is the north–south angular difference and $\Delta y$ is the east–west angular difference. Next, we calculate the intermediate term $a$, which corresponds to the squared half-chord length between the two points on the unit sphere:

% \vspace{-3mm}
\begin{equation}
a = \sin^2\!\left(\frac{\Delta x}{2}\right) + \cos(x_1) \cdot \cos(x_2) \cdot \sin^2\!\left(\frac{\Delta y}{2}\right)
\end{equation}

This term combines the north–south and east–west components while considering the Earth's curvature. We then obtain the central angle $c$ between the two points, and multiply by $R$ to get the great-circle distance $d$ in kilometers:

% \vspace{-3mm}
\begin{equation}
c = 2 \cdot \arcsin\!\left(\sqrt{a}\right),
\end{equation}
\begin{equation}
d = R \cdot c.
\end{equation}

Here, $d$ represents the shortest distance over the Earth's surface between the predicted and actual locations.Finally, the distance-based reward $r_{\text{distance}}$ is defined as a piecewise function of $d$, decreasing smoothly as $d$ increases and capped at 20000 km (the approximate maximum great-circle distance between two antipodal points, $\pi R \approx 20000$ km):

\begin{equation}
r_{\text{distance}} =
\begin{cases}
1.0 - 0.5 \cdot \frac{d}{750}, & d \leq 750, \\[4pt]
0.5 - 0.3 \cdot \frac{d - 750}{1750}, & 750 < d \leq 2500, \\[4pt]
0.2 - 0.2 \cdot \frac{d - 2500}{17500}, & \text{otherwise}.
\end{cases}
\end{equation}

This reward structure ensures high rewards for precise predictions, moderate penalties for mid-range errors, and gradual decay for very distant predictions, thereby maintaining informative gradient signals even for large errors.

\paragraph{Format-Based Reward}  
Models often exhibit \emph{format hallucinations}, producing outputs with missing coordinates, multiple ambiguous candidates, or inconsistent placement, which hinders downstream evaluation.  
To address this, we define a binary format reward that assigns $1$ only when the output contains exactly one valid decimal latitude–longitude pair in the expected form—enclosed in parentheses and separated by a comma; otherwise, it assigns $0$.  
This encourages strict adherence to the output template, penalizes ambiguous or ill-structured predictions, and, when combined with the distance-based reward, ensures contributions come only from accurate and well-formatted predictions.

% \vspace{-3mm}
\begin{equation}
\label{eq:format-reward}
r_{\text{format}} =
\begin{cases}
1, & \text{valid coordinate pair in expected format}, \\
0, & \text{otherwise}.
\end{cases}
\end{equation}

%\paragraph{Final Reward}
The overall reward $r$ for a single sample combines both components:

% \[
% r = r_{\text{distance}} * r_{\text{format}},
% \]

% \vspace{-5mm}
\begin{equation}
r = r_{\text{distance}} \times r_{\text{format}},
\end{equation}

% \[
% r = r_{\text{distance}} + r_{\text{trace}},
% \]

\noindent
where $r_{\text{distance}}$ captures the spatial precision of the prediction, and $r_{\text{format}}$ reflects the semantic alignment and interpretability of the reasoning chain. The total reward serves as the training signal in the reinforcement learning phase.

\subsection{Diverse Geographic Reasoning Data Selection}

% \paragraph{Diverse Data Selection}

To enhance the model’s generalization ability across diverse and geographically challenging scenarios, we construct a hard subset from the MP16-Pro~\cite{jia2024g3} dataset, denoted as MP16-Hard-200k. Specifically, we first identify 100k MP16-Pro~\cite{jia2024g3} samples that Qwen-VL-3B~\cite{Qwen2.5vl} localizes correctly; these samples form dense spatial clusters concentrated around major population centers and iconic landmarks—regions we term “popular regions” due to the model’s likely prior exposure or access to strong visual cues. To extract more challenging cases, we exclude all samples located within a 200-kilometer radius of any such popular region. The remaining samples, typically drawn from remote, visually ambiguous, or culturally neutral areas, compose the MP16-Hard-200k subset. These examples present greater localization difficulty and are used to augment model training, encouraging stronger spatial reasoning in underrepresented geographies. This targeted selection approach serves as a practical means of pushing models beyond surface-level pattern recognition, promoting robust geospatial understanding under real-world, long-tail conditions.

\begin{table*}[ht]
\centering
\small
\renewcommand{\arraystretch}{0.9}
\resizebox{\linewidth}{!}{
\begin{tabular}{ccccccc | ccccc} 
\midrule

\multicolumn{2}{c}{} & \multicolumn{5}{c}{IM2GPS3K} & \multicolumn{5}{c}{YFCC4K} \\

% \cline{3-12}
\cmidrule(lr){3-12}

\multicolumn{2}{c}{Methods} & 1km & 25km & 200km & 750km & 2500km & 1km & 25km & 200km & 750km & 2500km \\

\midrule
\multicolumn{4}{l}{\textit{Retrieval-based Methods}}  \\

[L]kNN, $\sigma=4$ & ICCV'17 & 7.2 & 19.4 & 26.9 & 38.9 & 55.9 & 2.3 & 5.7 & 11.0 & 23.5 & 42.0 \\
Img2Loc           & SIGIR'24 & 15.34 & 39.83 & 53.59 & 69.7 & 82.78 & 19.78 & 30.71 & 41.4 & 58.11 & 74.07 \\
PIGEON            & CVPR'24 & 11.3 & 36.7 & 53.8 & 72.4 & 85.3 & 10.4 & 23.7 & 40.6 & 62.2 & 77.7 \\
G3                & NeurIPS'24 & 16.65 & 40.94 & 55.56 & 71.24 & 84.68 & 23.99 & 35.89 & 46.98 & 64.26 & 78.15 \\
Geo-Ranker    & NeurIPS'25 & 18.79 & 45.05 & 61.49 & 76.31 & 89.29 & 32.94 & 43.54 & 54.32 & 69.79 & 82.45 \\

\midrule
\multicolumn{4}{l}{\textit{Retrieval-free Methods}}  \\
PlaNet            & ECCV'16 & 8.5 & 24.8 & 34.3 & 48.4 & 64.6 & 5.6 & 14.3 & 22.2 & 36.4 & 55.8 \\
CPlaNet           & ECCV'18 & 10.2 & 26.5 & 34.6 & 48.6 & 64.6 & 7.9 & 14.8 & 21.9 & 36.4 & 55.5 \\
ISNs              & ECCV'18 & 10.5 & 28.0 & 36.6 & 49.7 & 66.0 & 6.5 & 16.2 & 23.8 & 37.4 & 55.0 \\
Translocator      & ECCV'22 & 11.8 & 31.1 & 46.7 & 58.9 & 80.1 & 8.4 & 18.6 & 27.0 & 41.1 & 60.4 \\
GeoDecoder        & ICCV'23 & 12.8 & 33.5 & 45.9 & 61.0 & 76.1 & 10.3 & \textbf{24.4} & 33.9 & 50.0 & 68.7 \\
GeoCLIP           & NeurIPS'23 & 14.11 & 34.47 & 50.65 & 69.67 & 83.82 & 9.59 & 19.31 & 32.63 & 55.00 & 74.69 \\
GLOBE        & NeurIPS'25  & - & 40.18 & 56.19 & 71.45 & - & - & - & - & - & - \\
Geo-R(Ours)         &    & \textbf{18.10} & \textbf{41.53} & \textbf{58.31} & \textbf{75.33} & \textbf{86.42} & \textbf{10.47} & 22.67 & \textbf{40.04} & \textbf{60.83} & \textbf{75.84} \\

\hline
\end{tabular}}
\caption{Main results on IM2GPS3K and YFCC4K (higher is better). The best results achieved by retrieval-free methods are highlighted in bold. }
\label{tab:main_results}
% \vspace{-3mm}
\end{table*}

\section{Experiments}

\subsection{Datasets}

\paragraph{Training Data}
We primarily use two datasets—MP16-Rand-500K and MP16-Hard-200K—both constructed from the MP16-Pro dataset~\cite{jia2024g3} as described in the Methods section. MP16-Pro contains 4.72 million geotagged Flickr images annotated with hierarchical geographic labels, including country, region, city, and street level, providing a strong foundation for multi-level geographic reasoning. However, it also suffers from missing intermediate labels, inconsistent formatting, and multilingual place names, which reduce the quality of supervision. To address these issues, we standardize annotations by reverse-resolving GPS coordinates into canonical geographic entities, thereby completing missing labels and unifying diverse formats. Both datasets are compatible with the input-output formats required by SFT and RL.

\paragraph{Evaluation Benchmarks}
We evaluate our model on two widely used geolocation benchmarks to assess both fine-grained localization and global generalization.  The IM2GPS3K~\cite{hays2008im2gps} dataset consists of 3,000 manually curated Flickr images, each annotated with precise GPS coordinates. The images capture a wide variety of outdoor scenes—ranging from cityscapes and iconic landmarks to natural landscapes—making it a challenging benchmark for testing location prediction at high spatial resolution. Complementing this, the YFCC4K~\cite{thomee2016yfcc100m} dataset contains 4,000 geotagged images sampled from the broader YFCC100M corpus. To ensure balanced geographic coverage, the images are evenly distributed across continents and feature visually diverse content, including urban, rural, and natural settings. Compared to IM2GPS3K, YFCC4K places greater emphasis on testing cross-region generalization under globally heterogeneous environments.

\paragraph{Evaluation Metrics}
We follow standard protocols established in prior works~\cite{vivanco2023geoclip, zhou2024img2loc, jia2024g3}. Specifically, we compute the geodesic distance between the predicted and ground-truth coordinates for each test sample, and report the percentage of predictions that fall within predefined distance thresholds: 1km, 25km, 200km, 750km, and 2500km. This metric provides a holistic view of both near-exact localization and coarse regional accuracy.

\subsection{Baselines}

We compare our method against a range of representative baselines, which can be broadly categorized into retrieval-based and non-retrieval-based methods. Retrieval-based methods include [L]kNN~\cite{vo2017revisiting}, which aggregates coordinates from the top-k nearest images; Translocator~\cite{pramanick2022world}, a dual-branch transformer leveraging image and segmentation inputs; Img2Loc~\cite{zhou2024img2loc}, which incorporates retrieved coordinates into a RAG prompt; and PIGEON~\cite{haas2024pigeon}, which retrieves over semantic location clusters. In contrast, non-retrieval methods directly perform prediction without reference data. These include PlaNet~\cite{seo2018cplanet} and CPlaNet~\cite{seo2018cplanet}, which formulate localization as a classification task over geographical cells; ISNs~\cite{muller2018geolocation}, which fuses partition hierarchies and scene context; GeoDecoder~\cite{clark2023we}, which models hierarchical relationships via cross-attention; and GeoCLIP~\cite{vivanco2023geoclip}, which uses CLIP-based vision-language features with GPS embedding but no external image retrieval. 

% This taxonomy clarifies the methodological differences and highlights the retrieval-free nature of our proposed framework.

% \subsection{Experimental Setup}  
% \mf{moved here}

\subsection{Implementation Details}
We adopt a two-stage training strategy without relying on any retrieval module, and perform training on Qwen2.5-VL-7B-Instruct, a large vision-language model with built-in multi-modal reasoning capability. The model takes both image inputs and structured location prompts, and is trained to generate geographic reasoning chains along with hierarchical labels and final coordinates.

For the SFT Stage, we use a dataset of 500k samples consisting of images,  reasoning chains, and coordinate triples. We optimize a joint objective over coordinate regression and reasoning chain supervision using the AdamW~\cite{loshchilov2017decoupled} optimizer with a learning rate of $1 \times 10^{-5}$, batch size of 64, and a total of one epoch. 

% A weighting factor of $\lambda=0.7$ is applied between the two losses, and we adopt a hard-negative sampling strategy with $K^{(1)}=1$.

For the RL Stage, we perform geographic reasoning policy optimization using 200k samples. The model is optimized with policy gradients guided by the composite reward signal described in Section 3.2, incorporating both haversine distance and format. All experiments are conducted on a cluster of 8 NVIDIA A100 GPUs.

\subsection{Main Results}

Our experimental results underscore the effectiveness of combining structured reasoning with reward-guided optimization for image geolocation. As shown in Table~\ref{tab:main_results}, our method consistently outperforms all retrieval-free baselines and achieves performance comparable to retrieval-based approaches on the IM2GPS3K and YFCC4K benchmarks. On IM2GPS3K, our model achieves a 1km accuracy of 18.10\%, surpassing prior retrieval-free methods such as GeoCLIP (14.11\%) and GLOBE, and even approaching or exceeding several retrieval-based methods. At the 2500km level, our accuracy reaches 86.42\%, indicating a strong global localization capability. On the YFCC4K dataset, which contains more challenging and diverse images, our approach achieves 10.47\% at 1km and 75.84\% at 2500km, again outperforming all retrieval-free models and rivaling top retrieval-based systems. These results demonstrate the synergy between interpretable intermediate supervision and RL, highlighting the value of geographic reasoning as a core component in building robust, retrieval-free geolocation agents.

\begin{table}[t!]
\centering
\renewcommand{\arraystretch}{1.1}
\resizebox{\columnwidth}{!}{
\begin{tabular}{cc|ccccc}
\toprule
\textbf{Size} & \textbf{Data} & \textbf{1km} & \textbf{25km} & \textbf{200km} & \textbf{750km} & \textbf{2500km} \\
\midrule
3B & -   & 3.4\% & 15.9\% & 33.4\%  & 48.4\% & 61.4\% \\
3B & CoT & 3.7\%  & 16.4\%  & 47.5\%  & 48.2\%  & 66.7\%  \\
3B & CoR & 4.5\% & 24.1\% & 44.8\% & 56.7\% & 69.6\% \\
\midrule
7B & -   & 5.3\% & 24.3\% & 42.4\% & 61.4\% & 72.9\% \\
7B & CoT & 6.3\% & 26.1\% & 44.6\% & 60.6\% & 71.9\% \\
7B & CoR & 7.1\% & 33.7\% & 55.5\% & 73.4\% & 85.5\% \\
\midrule
32B & -   & 10.2\%  & 29.7\%  & 43.1\%  & 68.4\%  & 73.9\%  \\
32B & CoT & 11.0\% & 33.7\% & 50.5\% & 67.0\% & 82.5\% \\
32B & CoR & 12.3\% & 35.0\% & 50.7\% & 66.7\% & 81.4\% \\
% CLIP & - & 15.8\% & 33.9\% & 52.1\% & 71.7\% & 84.7\% \\
% \midrule
% 3B & 10k & 5.8\% & 25.6\% & 48.3\% & 67.7\% & 82.3\% \\  
% 3B & 100k & 8.6\% & 27.8\% & 47.6\% & 67.2\% & 81.5\% \\
% 3B & 200k & - & -- & -- & -- & --  \\
% 3B & 500k & 16.6\% & 35.7\% & 53.8\% & 71.2\% & 84.4\% \\

\bottomrule
\end{tabular}
}

\caption{
Comparison of geolocation accuracy across different model sizes and prompting strategies: baseline, CoT for Chain-of-Thought prompting, and CoR for Chain-of-Region reasoning. Accuracy is reported at five distance thresholds from 1km to 2500km on the IM2GPS3K benchmark. CoR introduces hierarchical geographic reasoning before final coordinate prediction.
}
\label{tab:geoloc_accuracy}
% \vspace{-3mm}
\end{table}

\subsection{Ablation Study}

To comprehensively analyze the factors that influence geolocation performance, we conduct ablation studies across four dimensions: reasoning strategy, reasoning supervision, data selection, and reinforcement learning. First, in Reasoning with CoR, we compare three prompting strategies—baseline, Chain-of-Thought (CoT), and our proposed CoR—to evaluate the impact of explicit reasoning guidance. Second, in Training with CoR, we examine how incorporating structured reasoning chains as supervision, under varying data scales, affects localization accuracy. Third, in Diverse Data Selection, we assess the effect of sampling strategies by comparing random sampling with hard-case sampling, focusing on both fine-grained accuracy and generalization. Finally, in Reinforcement Learning, we investigate how reward-guided optimization improves performance under different supervision settings, particularly in handling challenging examples. Together, these studies highlight the critical role of structured reasoning, data composition, and learning strategy in enhancing both the precision and robustness of geolocation models.  Due to the high computational cost of reinforcement learning, we conduct ablation studies using only the 3B parameter-scale model.

\paragraph{Reasoning with CoR}   We evaluate the geolocation capabilities of Qwen2.5-VL models across three parameter scales—3B, 7B, and 32B—under different prompting strategies, including the baseline, CoT, and our proposed CoR. As shown in Table~\ref{tab:geoloc_accuracy}, performance consistently improves with both model scale and the introduction of reasoning guidance. CoR significantly outperforms both the baseline and CoT across all distance thresholds. For example, the 7B model with CoR achieves 33.7\% at 25km and 85.5\% at 2500km, compared to 26.1\% and 71.9\% with CoT, demonstrating notable gains in both fine-grained and coarse-level localization. Similar improvements are observed with the 32B model. These results indicate that scaling alone is insufficient, and incorporating explicit geographic reasoning through CoR yields substantial benefits, particularly in the mid- to long-range prediction bands from 200km to 2500km.

\begin{table}[t!]
\centering
\renewcommand{\arraystretch}{1.1}
\resizebox{\columnwidth}{!}{
\begin{tabular}{ll|ccccc}
\toprule
\textbf{Size} & \textbf{Type} & \textbf{1km} & \textbf{25km} & \textbf{200km} & \textbf{750km} & \textbf{2500km} \\ 
\midrule
-- & -- & 4.5\% & 24.1\% & 44.8\% & 56.7\% & 69.6\% \\
\midrule
10k & - & 5.8\% & 25.6\% & 48.3\% & 67.7\% & 82.3\% \\ 
10k & CoR & 7.3\% & 28.1\% & 48.6\% & 69.0\% & 83.5\% \\
100k & - & 8.6\% & 27.8\% & 47.6\% & 67.2\% & 81.5\%\\
100k & CoR & 9.2\% & 29.9\% & 49.4\% & 69.2\% & 82.8\%\\
500k & -  & 10.4\% & 30.9\% & 48.6\% & 70.5\% & 83.7\% \\
500k & CoR & 12.6\% & 31.7\% & 50.2\% & 70.3\% & 84.3\% \\

\bottomrule
\end{tabular}
}
\caption{
Ablation study on the impact of data scale and prompt format using Qwen2.5-VL-3B. 
"Size" indicates the number of supervised SFT training samples. 
"Type" refers to the prompting format, where "-" uses a simple location-only prompt and "CoR" employs our proposed Chain-of-Region reasoning structure.
}
\label{tab:qwen3b_sft_res}
% \vspace{-3mm}
\end{table}

\paragraph{Training with CoR}
We investigate how incorporating structured geographic reasoning into supervision affects model performance. Using Qwen2.5-VL-3B, we compare location-only training with our CoR format under varying data scales. As shown in Table~\ref{tab:qwen3b_sft_res}, training with CoR reasoning label significantly boosts geolocation accuracy across all distance thresholds. For instance, at 1km, accuracy improves from 7.3\% to 12.6\% when scaling from 10k to 500k samples, while at 2500km, it rises from 82.8\% to 84.3\%. Notably, CoR outperforms location-only baselines at each scale, with consistent gains of 0.6--2.0 percentage points. These results show that reasoning chains improve interpretability and directly enhance coordinate prediction, especially with limited data.

\paragraph{Diverse Data Selection} 
As shown in Table~\ref{tab:qwen3b_sft_results}, across both training sizes, random sampling (Rand) consistently outperforms hard-case sampling (Hard), especially at fine-grained levels such as 1km. For example, under the 100k setting, Rand achieves 9.2\% accuracy at 1km, compared to only 6.0\% with Hard—yielding a 3.2 percentage point gap. This performance difference is largely attributed to the nature of hard samples, which often include noisy scenes, ambiguous context, or conflicting multimodal cues.

While such examples are useful for stress-testing model robustness, supervised fine-tuning tends to overfit these difficult instances, as it treats all ground-truth labels as equally reliable and lacks uncertainty modeling. In contrast, random sampling covers a broader and more diverse range of geographic regions and scene types, enabling the model to learn more robust and transferable spatial features.

\begin{table}[t!]
\centering
\resizebox{\columnwidth}{!}{
\begin{tabular}{ll|ccccc}
\toprule
\textbf{Size} & \textbf{Sample} & \textbf{1km} & \textbf{25km} & \textbf{200km} & \textbf{750km} & \textbf{2500km} \\ 
 
\midrule
-- & -- & 4.5\% & 24.1\% & 44.8\% & 56.7\% & 69.6\% \\
\midrule
10k & Rand & 7.3\% & 28.1\% & 48.6\% & 69.0\% & 83.5\% \\
10k & Hard & 4.9\% & 27.1\% & 47.8\% & 67.4\% & 81.1\% \\

% \midrule
100k & Rand & 9.2\% & 29.9\% & 49.4\% & 69.2\% & 82.8\%\\
100k & Hard & 6.0\% & 28.5\% & 48.9\% & 68.0\% & 82.3\% \\
% 500k & Rand & 12.6\% & 31.7\% & 50.2\% & 70.3\% & 84.3\% \\
\bottomrule
\end{tabular}
}
\caption{
Geolocation accuracy of Qwen2.5-VL-3B trained with supervised SFT under varying data sizes and sampling strategies. 
All experiments adopt the CoR prompt format unless marked with missing entries. 
"Size" indicates the number of training samples; "Type" denotes the data selection method, including Rand for random sampling and Hard for hard-case mining.
}
\label{tab:qwen3b_sft_results}
% \vspace{-3mm}
\end{table}

\paragraph{Reinforcement Learning}

We further investigate the role of RL in optimizing geolocation performance under different supervision conditions. As shown in Table~\ref{tab:qwen3b_r1_results}, RL brings consistent improvements, particularly when applied to hard-case data. For instance, under low-resource SFT (10k), applying RL on hard samples improves 1km accuracy from 7.3\% to 8.6\%, outperforming RL on random samples (6.7\%). This suggests that reward-guided optimization better exploits complex examples that are challenging for conventional supervised learning. At larger SFT scales (500k), RL continues to improve performance. Notably, hard-sample RL reaches 14.8\% at 1km and 83.8\% at 2500km, outperforming both the SFT-only baseline and RL with random data. The gains are especially pronounced in fine-grained localization. These results demonstrate that while random samples contribute to stable SFT training, hard samples become more valuable in RL settings, where reward signals help extract meaningful gradients and mitigate overfitting—ultimately improving generalization to complex, real-world geographic scenarios.

\subsection{Result Analysis}
Experimental results show that the CoR prompting strategy improves coordinate prediction by breaking down geolocation into structured, hierarchical reasoning steps, thereby aligning the generation process with the model’s internal representations and reasoning mechanisms.The use of structured reasoning data during training further reinforces this capability. However, SFT remains insensitive to coordinate-level errors, making it difficult to optimize localization accuracy precisely~\cite{chu2025sft,shen2025vlm}. With the introduction of RL, the model can directly optimize against geodesic distance, leading to significant improvements in precision—particularly in complex or geographically ambiguous regions.

\begin{table}[t!]
\centering
\resizebox{\columnwidth}{!}{
\begin{tabular}{lc|ccccc}
\toprule

\textbf{Methods} & \textbf{Sample} & \textbf{1km} & \textbf{25km} & \textbf{200km} & \textbf{750km} & \textbf{2500km} \\  
\midrule
% -         & Zero  & 3.4\% & 15.9\% & 33.4\% & - & -\\
\multicolumn{7}{l}{\textit{\textbf{Without} Chain of Region}} \\
SFT-10k        & - & 5.8\% & 25.6\% & 48.3\% & 67.7\% & 82.3\% \\
+RL         & Rand 10k   & 7.0\% & 27.6\% & 41.8\% & 65.5\% & 80.6\% \\
+RL         & Hard 10k  & 7.8\% & 28.6\% & 52.7\% & 66.9\% & 80.7\% \\

\midrule
\multicolumn{7}{l}{\textit{With Chain of Region}} \\

SFT-10k        & - & 7.3\% & 28.1\% & 48.6\% & 69.0\% & 83.5\% \\
+RL         & Rand 10k  & 6.7\% & 24.3\% & 42.5\% & 61.0\% & 74.9\% \\
+RL         & Hard 10k  & 8.6\% & 28.7\% & 45.7\% & 69.0\% & 85.4\% \\

\midrule
\multicolumn{7}{l}{\textit{With Chain of Region}} \\
SFT-500k         & -  & 12.6\% & 31.7\% & 50.2\% & 70.3\% & 84.3\% \\
% \midrule
+RL   & Rand 10k & 11.3\%  & 28.9\%  & 49.6\%  & 68.7\%  & 79.3\%  \\
+RL   & Hard 10k  & 11.1\% & 31.6\% & 50.3\% & 66.2\% & 78.3\% \\

\midrule
\multicolumn{7}{l}{\textit{With Chain of Region}} \\
SFT-500k         & -  & 12.6\% & 31.7\% & 50.2\% & 70.3\% & 84.3\% \\
% \midrule
+RL   & Rand 200k & 13.3\% & 32.4\% & 51.6\% & 71.3\% & 82.3\% \\
+RL   & Hard 200k & 14.8\% & 36.3\% & 54.6\% & 72.7\% & 83.8\% \\

\bottomrule
\end{tabular}
}
\caption{
Evaluation of Qwen2.5-VL-3B under RL with different SFT scales and reward strategies. “SFT-10k/500k” indicates the number of samples used for SFT. “Rand” and “Hard” refer to randomly sampled or hard-case data used in RL. Results demonstrate the effect of SFT and reward-guided RL across multiple geodesic thresholds.}
\label{tab:qwen3b_r1_results}
% \vspace{-3mm}
\end{table}

 \section{Conclusion}

In this paper, we introduced Geo-R, a retrieval-free and reasoning-driven framework for global image geolocalization that combines structured geographic reasoning with reinforcement learning based on verifiable, spatially grounded rewards. Our approach integrates a scalable Chain-of-Region paradigm for interpretable multi-level inference with reward-guided optimization that directly targets geographic plausibility. The results demonstrate that large language model reasoning, without any reliance on external retrieval or synthetic teacher supervision, can deliver competitive or superior localization accuracy, interpretability, and cross-domain generalization. Comprehensive experiments on standard benchmarks validate the effectiveness and robustness of Geo-R, underscoring the potential of reasoning-centric methods for building scalable and general-purpose geographic agents. We hope this work encourages further research into interpretable and verifiable multimodal reasoning, as well as new applications at the intersection of vision, language, and geospatial artificial intelligence.

\bibliography{aaai2026}

\newpage
\appendix

\bigskip
% \noindent Thank you for reading these instructions carefully. We look forward to receiving your electronic files!

\end{document}